\pdfoutput=1

\documentclass[11pt]{article}

\usepackage{acl}
\usepackage{amsmath}
\usepackage{times}
\usepackage{latexsym}

\usepackage[T1]{fontenc}

\usepackage[utf8]{inputenc}

\usepackage{microtype}

\usepackage{inconsolata}

\usepackage{graphicx}

\title{Advancing Annotation of Stance in Social Media Posts: A Comparative Analysis of Large Language Models and Crowd Sourcing}

\author{Mao Li \\
  University of Michigan   \\Ann Arbor, USA\\
  \texttt{maolee@umich.edu} \\\And
  Frederick Conrad\\
   University of Michigan  \\ Ann Arbor, USA \\
  \texttt{fconrad@umich.edu} \\}

\begin{document}
\maketitle
\begin{abstract}

In the rapidly evolving landscape of Natural Language Processing (NLP), the use of Large Language Models (LLMs) for automated text annotation in social media posts has garnered significant interest. Despite the impressive innovations in developing LLMs like ChatGPT, their efficacy, and accuracy as annotation tools are not well understood. In this paper, we analyze the performance of eight open-source and proprietary LLMs for annotating the stance expressed in social media posts, benchmarking their performance against human annotators’ (i.e., crowd-sourced) judgments. Additionally, we investigate the conditions under which LLMs are likely to disagree with human judgment. A significant finding of our study is that the explicitness of text expressing a stance plays a critical role in how faithfully LLMs’ stance judgments match humans’. We argue that LLMs perform well when human annotators do, and when LLMs fail, it often corresponds to situations in which human annotators struggle to reach an agreement. We conclude with recommendations for a comprehensive approach that combines the precision of human expertise with the scalability of LLM predictions. This study highlights the importance of improving the accuracy and comprehensiveness of automated stance detection, aiming to advance these technologies for more efficient and unbiased analysis of social media.
\end{abstract}

\section{Introduction}
A primary challenge in automated social media analysis is translating unstructured textual data, i.e., users' posts, into a structured format, such as categories or numerical values (\citealp{oconnor_tweets_2010}; \citealp{conrad2021social}) to correctly interpret what a post is saying, so that the content can be quantitatively analyzed. Stance detection has emerged as a promising solution to this challenge, offering an automatic approach to classifying the opinion or position that the user has expressed in a post, typically one of several options such as favoring, opposing, or being neutral with respect to the topic, or being irrelevant \citep{kucuk_tutorial_2022}. This method has become increasingly common in social media analysis for conducting social research, i.e., extracting public opinion from social media postings (\citealp{aldayel2021stance}; \citealp{karande2021stance}; \citealp{bode2020study}).  

Historically, stance detection has used traditional machine learning techniques such as Support Vector Machines (SVM) and logistic regression, alongside deep learning methods like Convolutional Neural Networks (CNN) and Long Short-Term Memory networks (LSTM) (\citealp{kuccuk2020stance}; \citealp{kucuk_tutorial_2022}). More recently, researchers have made significant strides with the introduction of pre-trained language models such as BERT, which have substantially improved prediction accuracy (\citealp{karande2021stance}). However, the requirement for high-quality annotation data remains critical. The most prevalent approach to annotating stance involves crowd-sourcing on platforms like Amazon Mechanical Turk. Yet, the process can be both time-consuming and expensive. In response to these challenges, researchers have begun to explore stance prediction based on only a few or even no training examples, i.e., Few- or Zero-Shot learning  (\citealp{allaway_zero-shot_2020}; \citealp{allaway_zero-shot_2023}; \citealp{alturayeif_systematic_2023}; \citealp{darwish2020unsupervised}).

The OpenAI team's introduction of the concept that language models can act as Few-Shot learners (\citealp{brown2020language}) highlights the remarkable ability of these large language models (LLMs) to understand complex language. This development opens new possibilities for using LLMs to accurately extract or infer users’ opinions and knowledge from their posts. 

Building upon the established capabilities of ChatGPT, recent research has investigated the application of LLMs in annotating stances of social media texts (\citealp{gilardi_chatgpt_2023}; \citealp{zhang_how_2023}). These studies demonstrate that ChatGPT attains state-of-the-art performance in multiple stance detection benchmarks (\citealp{zhang_how_2023}) and, in certain annotation tasks, surpasses human annotators, as validated by expert assessments (\citealp{gilardi_chatgpt_2023}). \citet{10.1162/coli_a_00502} further argues that LLMs could revolutionize computational social science by serving as efficient Zero-Shot data annotators within human annotation teams, potentially transforming the approach to stance detection tasks.

However, these findings and resultant conclusions warrant further examination. \citet{aiyappa_can_2023} highlight potential data contamination in ChatGPT's evaluations on widely used stance detection benchmarks, casting doubt on their validity. Furthermore, \citet{cruickshank_use_2023} indicate that Few- or Zero-Shot learning with LLMs may not consistently outperform supervised methods in certain datasets. A broader evaluation by \citet{kocon_chatgpt_2023} reveals that ChatGPT does not universally outperform the SOTA model across all tasks. Corroborating this, research indicates that LLM performance is task- or dataset-specific \cite{zhu2023chatgpt}. Given the diversity of LLMs currently available, a comprehensive understanding of when and why LLMs excel in stance annotation tasks, and which specific tasks they are best suited for, remains elusive.

One factor contributing to the inconsistent results of LLMs in stance annotation tasks may be their varied ability to process texts with different degrees of implicitness. Stance detection involves interpreting not just overtly stated information but also nuances and implications within the text \citep{kucuk_tutorial_2022}, making the clarity with which a stance is expressed in the text a critical factor for accurate annotation. While identifying explicitly stated stances is relatively straightforward, inferring those that are implied presents a greater challenge. In the context of social media, where the benefit of stance detection is potentially great, Kahneman's distinction between fast, i.e., automatic and intuitive thinking (System 1), and slow, i.e., analytical and deliberate (System 2) thinking, may be quite relevant: ``the form, speed, and images of digital content on social media are more likely to trigger `fast thinking,' by which intuitive and emotional biases can be exacerbated'' \citep{kahneman2011thinking}. This implies that recognizing explicitly stated stances, such as opinions, is generally more intuitive than analytic. However, social media also includes posts that require deeper, more critical thinking, where readers must delve into the underlying meanings. Whether understanding a particular post requires fast or slow thinking may depend on how explicitly the stance is expressed in the post, and this, in turn, is likely to affect LLMs' ability to annotate the stance correctly.

One aspect that influences how directly users express their opinion (stance) is the sensitivity or controversy of the topic. More controversial topics tend to lead to more explicit expressions of stance due to the emotional engagement or contentious nature of these subjects. For instance, in online discussions about the Supreme Court's ruling in Dobbs v. Jackson Women's Health Organization, posts often present clear and unambiguous stances – either favoring or opposing the Supreme Court’s decision. This pattern highlights how the sensitivity or controversy of a topic can significantly influence the clarity and directness with which individuals communicate their positions. On the flip side, when the goal is to measure knowledge-based awareness—for example, assessing how familiar social media users are with the Dobbs decision—they are less likely to express their knowledge explicitly, focusing instead on emotional content. Societal norms and conventions, such as the expectation to avoid discussing controversial social issues or the taboo against expressing anti-immigrant sentiments, might lead individuals to share their opinions in a more vague or indirect manner. For instance, individuals who hold pro-white or anti-immigrant views might choose to voice their opinions through subtle comments or coded language rather than explicitly stating their positions. This affects how explicitly people communicate on social media platforms. Thus, while seeking to understand public awareness or opinions on sensitive issues, researchers and observers must navigate both the emotional intensity of responses and the potential reticence prompted by societal expectations.

In summary, the effectiveness of LLMs in stance annotation tasks depends on how explicitly the stance is expressed in the posts, which is, in turn, influenced by topic sensitivity and societal norms. Understanding these factors is essential for improving the accuracy and consistency of stance detection in social media.
Therefore, in this study, we distinguish the annotation tasks that require more superficial inference and those requiring deeper and more complicated inference, calling the first ``System 1'' and the second ``System 2'' tasks \citep{kahneman2011thinking}. System 1 tasks are those that can be done fast and intuitively primarily because the text explicitly states the stance, while System 2 tasks require analytical reasoning because the position expressed in the text is less explicit. For stance annotation, a System 1 task mostly involves extracting the information that is explicitly expressed in the context. A System 2 task involves inferring what the user had in mind from what is implied in the text, thus requiring additional mental operations. We propose that, when they are prompted to classify stance, LLMs will be more likely to agree with human coders' judgments when the stance is explicitly stated in the text, i.e., in a System 1 task, than when the stance is implicitly expressed, i.e., a System 2 task. 

Moreover, we hypothesize that the explicitness with which stance is expressed varies both within individual texts and across the entire corpus. As previously suggested, certain topics are more likely to elicit posts that clearly articulate opinions. For example, discussions about abortion rights often feature explicitly stated opinions, while topics related to race and gender might result in more nuanced expressions of stance. Additionally, even within a single topic, texts representing different viewpoints can exhibit varying degrees of explicitness. For instance, \citet{klar2016social} show that during the 2016 presidential election, Trump supporters tended to understate their support due to social desirability concerns, which might lead them to express their support more subtly on social media platforms. This phenomenon underscores the multifaceted nature of explicitness, influenced by both the thematic context and the nuanced perspectives within the corpus.

In this study, we will first benchmark several open-source and propitiatory LLMs in terms of their performance on stance annotation tasks. Additionally, for the leading models (open-source and proprietary), we seek to measure the disparity between human annotators’ and LLMs’ judgments and test our hypothesis of whether the performance/agreement between human coders and LLMs is consistently higher in a System 1 task than in the System 2 task. Our hypotheses are as follows: 


\begin{enumerate}
    \item  System 1 texts, characterized by explicit information which can be processed quickly and intuitively, are likely to yield higher agreement between LLM and human annotations than System 2 texts, which require slow, analytical processing due to their implicit expression of stance.
    \item  The likelihood of annotating a text categorized as a System 1 or System 2 task is influenced by its related topic and the expressed opinion. Certain topics and opinions inherently produce more texts that require analytical, slow processing (System 2) due to their implicit nature. This categorization is typically done by researchers or annotators who assess the cognitive demands of understanding the text.
    \item LLMs demonstrate varying degrees of performance in stance detection tasks across different corpora, and this performance disparity is influenced by whether the texts predominantly require System 1 or System 2 processing.
\end{enumerate}

\section{Methodology}
\subsection{Data Sets}
In this study, we curated a dataset of tweets concerning the proposed and controversial inclusion of a citizenship question in the 2020 US Census. This dataset was derived from a larger collection of approximately 3.5 million tweets, posted between January 1 and September 17, 2020, which mentioned the 2020 Census, the US Census Bureau, or one of its surveys. We extracted a subset of tweets from this corpus, which contained the keywords ``citizenship'' or ``citizen,'' yielding about 17,000 unique tweets pertinent to the citizenship question debate.

Our aim was to ascertain users’ knowledge that a citizenship question had been barred from inclusion in the Census 2020 by the US Supreme Court and to gauge their stance towards the inclusion of such a question. We sampled 1,000 tweets from a corpus of ``citizenship'' tweets for human annotation. So that the task would not require more effort and engagement than was realistic and fair to expect from human judges, we divided the 1000 tweets into 10 random subsets of 100; for each subset, we asked 10 Mechanical Turk workers to judge (1) the likelihood that the author of the tweet believed the census questionnaire would ask about citizenship, and (2) the extent to which the author favored or opposed asking a citizenship question on the 2020 Census. They registered their judgments by moving a slider to a point between -50 and 50, where -50 corresponded to ``Very Unlikely/Oppose'' and 50 to ``Very Likely/Favor.'' This produced 10 knowledge (likelihood) and 10 opinion (favor-oppose) judgments for each tweet from each set of ten judges for each of the 1000 tweets. The judgments were elicited for each group of ten judges during May and June 2022. We used 10 judges on the assumption that a relatively large number of judgments for each tweet would stabilize the average judgments and dampen the effect of outliers. By adopting a continuous scoring system (-50 to 50) over the commonly used categorical judgment, we can more accurately gauge users’ confidence levels and quantify their consensus or diversity through the standard deviation of their numerical judgments. The distribution of judgments, including mean values and standard deviations, are presented in Table 1.

\begin{figure*}
    \centering
    \includegraphics[width=0.8\textwidth]{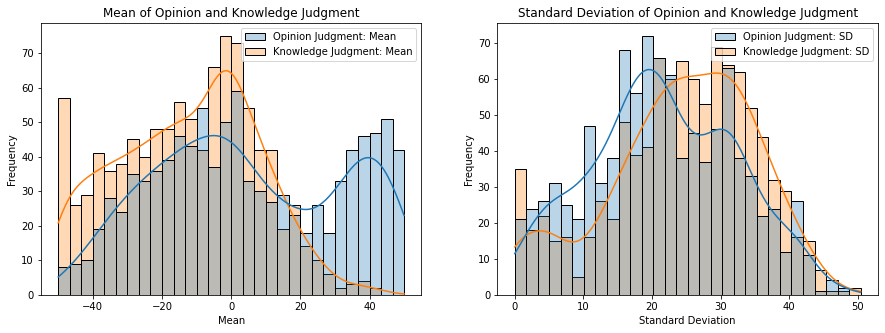}
    \caption{Distributions of mean and standard deviation of human judgments}
    \label{fig:dis}
\end{figure*}

To generate labels for training our predictive model, we averaged the ten likelihood judgments for each tweet. Furthermore, We categorized average judgments between -17 and -50 as ``Very Unlikely'' for the knowledge question, i.e., the author knew that the citizenship question had been barred from the Census, and ``Oppose'' for the opinion question, i.e., that the author of the post was opposed to including a citizenship question on the 2020 Census. Judgments from 17 to 50 were classified as ``Very Likely,'' i.e., that the author did not know the citizenship question had been barred from the Census, for the knowledge question and ``Favor'' for the opinion questions, i.e., the author of the tweet favored including a citizenship question. Mean judges between -17 and 17 were considered ``Can’t tell'' for knowledge and ``Neutral'' for opinion. For clearer labeling, in the subsequent sections, we will use ``Aware'' instead of ``Very Unlikely'' to denote users who knew that the citizenship question had been barred from the Census, and ``Not Aware'', rather than ``Very Likely'', to indicate users who did not know the citizenship question had been barred from the Census. From the annotated data, we randomly selected 20 posts from each of the 10 subsets (total 200) for the test dataset used in our initial experiment comparing multiple LLMs.

\subsection{Selected Models}
In this study, we evaluated four open-source LLMs: Llama2-13b-chat and Llama2-70b-chat \citep{touvron2023llama}, Llama3-8b-Instruct and Llama3-70b-Instruct \citep{llama3modelcard}, Zephyr-7b-beta \citep{tunstall2023zephyr}, and Falcon-180b-chat \citep{refinedweb}. In addition, we analyzed the performance of two proprietary models, gpt-3.5-turbo (0301 version) and gpt-4 (1106 version).

\subsection{Hardware}
Our experiments were carried out on a high-performance computing cluster, configured with Redhat8, Intel Xeon Gold 6226R CPUs, 180 GB of RAM, and five NVIDIA A40 GPUs, each with 48 GB of VRAM.

\subsection{Prompt}
To assess the LLMs' capabilities, we prepared two distinct prompting templates, i.e., Few-Shot setting and Zero-Shot setting (see Appendix~\ref{sec:appendix}), drawing on existing research, such as \citealp{zhao_calibrate_2021}; \citealp{liu_pre-train_2021} and \citealp{brown2020language}. The first template was straightforward, simply outlining the task, while the second included two corpus examples with annotations from our study team for reference. 


\subsection{Statistical Analysis}
To benchmark the performance of LLMs, we conducted evaluations using two distinct datasets: knowledge judgments and opinion judgments — sourced from the citizenship corpus. These evaluations examined two different prompt templates to facilitate a comprehensive analysis. Detailed findings are articulated in Table~\ref{tab:llmCompare}, which we will explore in subsequent discussions. For a deeper dive, we selected Llama3-70b-Instruct and GPT-4, which represent the forefront of open-source and proprietary LLM technologies, respectively. Our goal was to pinpoint the circumstances under which their judgments diverge from human annotations. To this end, we introduced a binary indicator variable to capture agreement between LLMs and human annotators (1 indicating agreement, 0 indicating disagreement).

In this investigation, we sought to test our three hypotheses empirically using the citizenship dataset. It's essential to highlight that our analytical framework employs the notion of ``explicitness'' within our research questions and hypotheses. However, quantifying explicitness is complex and subjective. To address this, we adopted the standard deviation of judges' ratings—reflecting the extent of disagreement among the MTurkers—as an indication of text explicitness. Echoing the insights of \citet{aroyo2015truth}, we believe that judges' disagreement is not inherently detrimental; rather, it can provide rich information. We argue that a large standard deviation, suggesting less consensus among the ten coders, likely stems from the nuanced articulation of stances within the text, which suggests a lower level of explicitness. In essence, we interpret judges' disagreement as a way to gauge the explicitness of the text.

To further explore this, with ten judges assigned to each annotation group and each judge responsible for annotating 100 tweets, we adopted a binary approach to gauge model alignment with human annotations—specifically, whether they agree or disagree. 

To investigate the first hypothesis (\textbf{H1}), we used logistic regression analysis to examine the relationship between agreement (agree/disagree) and the standard deviation associated with each question, as illustrated in Equation~\ref{eq2}, where $u_j$ represents the random intercept for the judge group $j$, capturing the variation in the log odds of agreement that is due to differences between groups not explained by the predictors in the model. The rationale here is that standard deviation serves as a proxy for explicitness; thus, we aimed to determine whether a higher level of explicitness (lower standard deviation of judgments) correlates with an increased likelihood of human annotator agreement with LLMs. In this analysis, we also added random intercepts across coder groups to mitigate any potential variability stemming from differences across annotation teams. This approach is predicated on the assumption that, although the standard deviation could reflect issues with annotation quality, by consistently assigning groups of ten coders to evaluate 100 tweets each and controlling for group variation, we can more reliably attribute observed variance to the content of the texts themselves rather than to disparities in annotation quality or groups. Note that we will not present the random intercept results since the estimates of random intercepts are not relevant to the point of the study. 

\begin{equation}
\text{logit}(P(\text{Agreement})) = \beta_0 + u_{j} + \beta_1\text{SD}
\label{eq2}
\end{equation}

To evaluate \textbf{H2}, our approach involves analyzing the average standard deviation (SD) across distinct corpora, segmented by task type and stance orientation. We employ pairwise t-tests to compare these corpora. The identification of significant differences will enable us to discern if the task and stance domain contribute to the variance in text explicitness.

To evaluate \textbf{H3}, we employed a bootstrapping methodology to analyze the F1 scores for the stances 'Favor/Aware' and 'Oppose/Not Aware' within each model, using human coders' evaluations as the truth. This procedure is predicated on the empirical support for \textbf{H2}. Assuming \textbf{H2} is confirmed, we aim to ascertain whether the hypothesis—that LLMs achieve greater accuracy on System 1 than System 2 tasks—applies not only at the level of individual texts but also across entire corpora, thereby extending our proposed mechanism from micro-level text analysis to macro-level corpus assessment.

\section{Results}
\subsection{Performance of LLMs}
\begin{table}[]
  \centering
  \caption{LLMs performance comparison}
  \label{tab:llmCompare}
  \resizebox{\columnwidth}{!}{%
  \begin{tabular}{|c|c|c|}
    \hline
    & \textbf{Citizenship-Opinion} & \textbf{Citizenship-Knowledge} \\

    Zero-Shot &  F1-Favor/F1-Oppose & F1-Not Aware/F1-Aware \\
    \hline\hline
    \textbf{Zephyr-7b-beta} &  0.75/0.15 & 0.06/0.26 \\
    \textbf{Llama3-8b-Instruct} &  0.22/0.46 & 0.26/0.63 \\
    \textbf{Llama2-13b-chat} &  0.03/0.23 & 0.05/0.17 \\
    \textbf{Llama2-70b-chat} &  0.39/0.36 & 0.06/0.61 \\
    \textbf{Llama3-70b-Instruct} &  0.83/0.58 & 0.11/0.60 \\
    \textbf{Falcon-180b-chat} & 0.51/0.06 & 0.12/0.56 \\
    \hline
    \textbf{gpt-3.5-turbo} &  0.28/0.30 & 0.00/0.64 \\
    \textbf{gpt-4}  & 0.82/0.58 & 0.00/0.65 \\

    \hline\hline
    Few-Shot & &    \\
    \hline\hline
    \textbf{Zephyr-7b-beta}  & 0.79/0.16 & 0.30/0.60 \\
    \textbf{Llama3-8b-Instruct} &  0.61/0.50 & 0.31/0.62 \\
    \textbf{Llama2-13b-chat} & 0.28/0.20 & 0.22/0.22 \\
    \textbf{Llama2-70b-chat}  & 0.60/0.42 & 0.25/0.59 \\
    \textbf{Llama3-70b-Instruct} &  \textbf{0.90/0.72} & \textbf{0.37/0.65} \\
    \textbf{Falcon-180b}  & 0.71/0.31 & 0.30/0.53 \\
    \hline
    \textbf{gpt-3.5-turbo}  & 0.68/0.30 & 0.24/0.67 \\
    \textbf{gpt-4}  & \textbf{0.86/0.61} & \textbf{0.35/0.80} \\
    \hline
  \end{tabular}
  }
\end{table}

In this section, we benchmark the performance of various LLMs to evaluate their effectiveness in stance detection tasks. Table 1 displays the F1 scores for each LLM under two settings: Zero-Shot and Few-Shot. For each setting, the first six large language models (LLMs) are open-source, whereas the last two are proprietary. The highest average F1 scores for both open-source and proprietary models are highlighted in bold to indicate the highest performance.

In proprietary LLMs, gpt-3.5-turbo exhibited a considerable performance gap compared to gpt-4. Comparing leading models from both open-source and proprietary categories (highlighted in bold), it appears that Llama3-70b-Instruct is competitive with gpt-4, the state-of-the-art proprietary LLM, especially in the Few-Shot setting.

Moreover, comparing the F1 scores for the same model across the table, we see that Few-Shot prompting significantly enhances the LLMs' performance of the annotation task compared to Zero-Shot. This is evident from the higher F1 scores, which suggest that providing a few examples helps models better interpret what they are being instructed (prompted) to do. This enhanced performance highlights the importance of context and examples in training LLMs for complex tasks.

We also observed that the reasons provided by LLMs for their difficulty classifying stance in a post reflect limitations in their reasoning ability compared to human annotators. For example, although human annotators were able to classify the user’s stance (i.e., their knowledge) about the Surpreme Court’s decision to bar a citizenship quesiotn fomr the 2020 Census, Llama2-13b-chat responded with ``Can’t tell'' citing a lack of explicit information about the user’s awareness. 

\subsection{\textbf{H1.} Agreement Between LLM and Human Annotations in System 1 vs. System 2 Texts}
\begin{table}[h]
\centering
\resizebox{\columnwidth}{!}{
\begin{tabular}{@{\extracolsep{5pt}}lcccc}
\\[-1.8ex]\hline
\hline \\[-1.8ex]
& \multicolumn{4}{c}{\textit{Dependent variable: agreement}} \\
\cline{2-5}
\\[-1.8ex] & (1) & (2) & (3) & (4) \\
\hline \\[-1.8ex]
Intercept & 1.239$^{***}$ & 0.836$^{***}$ & 1.511$^{***}$ & 0.048 \\
& (0.244) & (0.241) & (0.256) & (0.279) \\
SD\(_{\text{opinion}}\) & -0.044$^{***}$ & & -0.043$^{***}$ & \\
& (0.007) & & (0.007) & \\
SD\(_{\text{knowledge}}\) & & -0.058$^{***}$ & & -0.057$^{***}$ \\
& & (0.007) & & (0.008) \\
\hline \\[-1.8ex]
Observations & 1000 & 1000 & 1000 & 800 \\
Pseudo $R^2$ & 0.042 & 0.064 & 0.041 & 0.063 \\
\hline
\hline \\[-1.8ex]
\textit{Note:} & \multicolumn{4}{r}{$^{*}$p$<$0.1; $^{**}$p$<$0.05; $^{***}$p$<$0.01} \\
\end{tabular}
}
\caption{Logistic regression results}
\label{tab:logistic_regression}
\end{table}

The results from Equation  \ref{eq2}, presented in Table \ref{tab:logistic_regression}, establish a noteworthy relationship: variability among human judgments (i.e., higher standard deviation) is negatively correlated with agreement between LLMs and human judges. Our H1 is supported, indicating that LLMs, like humans, struggle with the interpretation of ambiguous textual content. This trend (i.e., as the variation in human coders’ responses increased, the chances that LLMs would agree with them went down) was consistently observed across all four logistic models

\subsection{\textbf{H2.} Influence of Topic and Opinion on Text Categorization as System 1 or 2}
\begin{figure}{}
        \centering
        \includegraphics[width=0.45\textwidth]{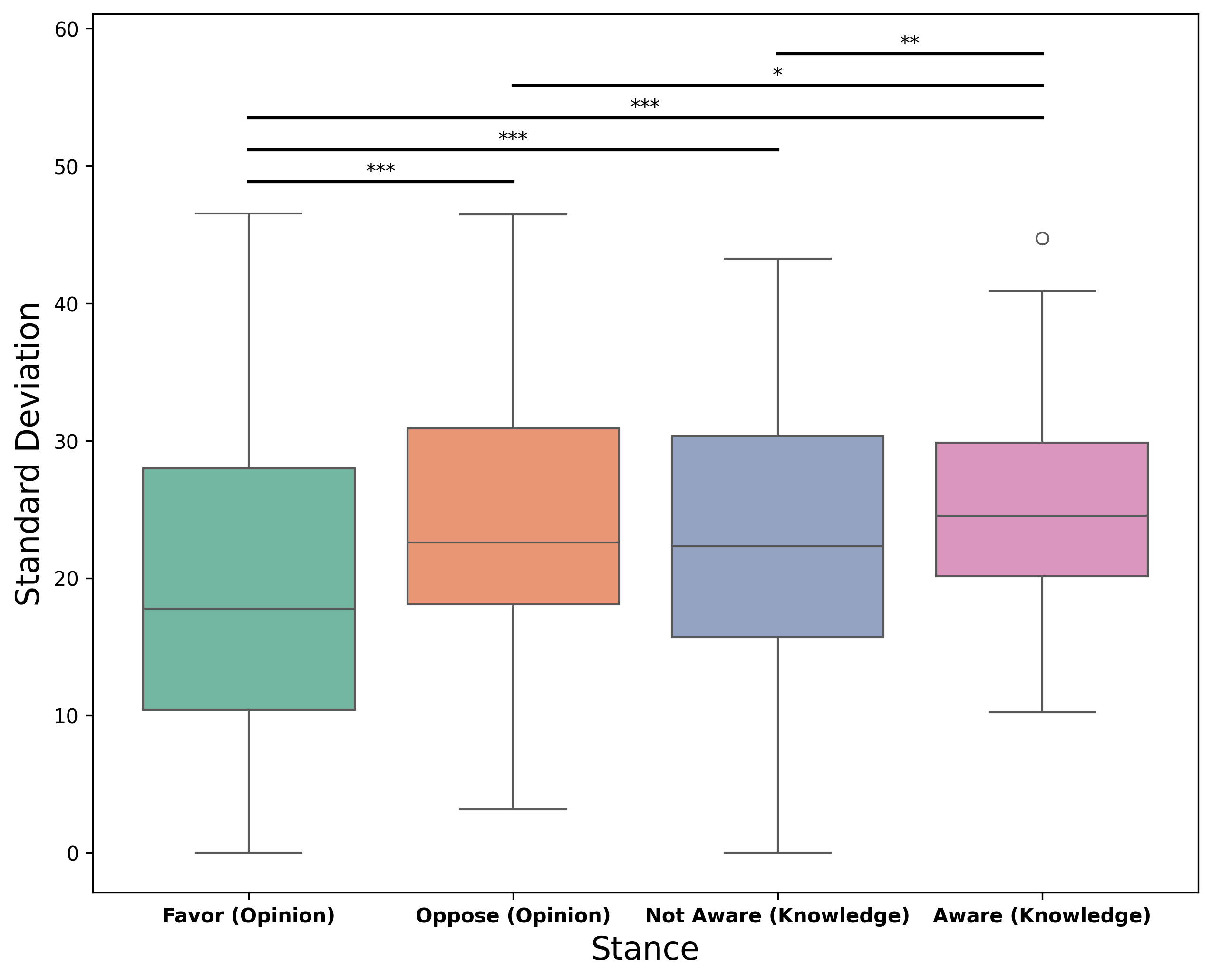}
        \caption{Box Plot and Pairwise T-Test for different scenarios}
        \label{fig:task_stance}
\end{figure}

In our \textbf{H2}, we proposed that the explicitness of texts is not just an individual attribute but could also be related to the topic and stance domain. In order to elucidate the complexities inherent in annotating stances, we created a taxonomy of what is being annotated, ``knowledge'' or ``opinion'' and how explicitly the stance is expressed (standard deviation of human judgments).  Each cell in Table~\ref{tab:task_stance} represents a different combination of these factors, which we refer to as ``scenarios''. This categorization distinguishes between tasks associated with `System 1' and `System 2' cognitive processes, as well as stances that are associated with System 1 or System 2 reasoning. Furthermore, we conducted a pairwise t-test between different scenarios as shown in Figure~\ref{fig:task_stance}. Based on the pairwise t-test, only Oppose(Opinion) and Not Aware (Knowledge) have no significant difference. This finding suggests that the scenarios represent different points on a spectrum, in terms of explicitness: at one end, the most straightforward combination `System 1' task with a `System 1' stance (upper left cell in the table), and at the other of the continuum, the more inferentially challenging pairing of a `System 2' task with a `System 2' stance (lower right cell in the table).  Stance detection in the off-diagonal cells is of intermediate difficulty. 
\begin{table}[]
        \centering
        \begin{tabular}{|l|c|c|}
            \hline
            & \begin{tabular}[c]{@{}c@{}}\textbf{System 1}\\ \textbf{Task}\end{tabular} & \begin{tabular}[c]{@{}c@{}}\textbf{System 2}\\ \textbf{Task}\end{tabular} \\ \hline
            \begin{tabular}[c]{@{}l@{}}\textbf{System 1}\\ \textbf{Stance}\end{tabular} & \begin{tabular}[c]{@{}c@{}}Favor stance\\ (Opinion) \\ mean SD: 19.81\end{tabular} & \begin{tabular}[c]{@{}c@{}}Aware stance\\ (Knowledge) \\ mean SD: 21.90\end{tabular} \\ \hline
            \begin{tabular}[c]{@{}l@{}}\textbf{System 2}\\ \textbf{Stance}\end{tabular} & \begin{tabular}[c]{@{}c@{}}Oppose stance\\ (Opinion) \\ mean SD: 22.68\end{tabular} & \begin{tabular}[c]{@{}c@{}}Not Aware stance\\ (Knowledge) \\ mean SD: 25.31\end{tabular} \\ \hline
        \end{tabular}
        \caption{Scenarios about different types of tasks and stances}
        \label{tab:task_stance}
\end{table}

Results from Table~\ref{tab:task_stance} and Figure~\ref{fig:task_stance} indicate that annotation complexity also varies significantly within each category of knowledge and opinion, which also aligns with our \textbf{H2}. Explicit acknowledgments that the Census will not include a citizenship question are easier than more nuanced expressions of unawareness, often interwoven with political discourse. For example, a person who was unaware that the Census would not include a citizenship question tweeted, ``This is great! Undocumented citizens and illegal aliens should not be counted towards the census!'' The expression ``This is great'' reflects a positive reaction to the Trump administration's effort to exclude undocumented immigrants from the Census. This tweet illustrates the user's lack of awareness of the Supreme Court's decision not to include the citizenship question vaguely. Similarly, explicit endorsements (i.e., an opinion rather than knowledge) of the citizenship question's inclusion are easier than more implicit expressions of opposition. Hence, it appears that various topics and stance domains are likely to produce texts with differing levels of explicitness.

\subsection{\textbf{H3.} LLM Performance Variations in Stance Detection Across Different Corpora}

Our goal remains to connect corpus-level explicitness with the performance of LLMs. Our analyses confirm that tasks and stances associated with `System 1' tend to be straightforward and facilitate the direct extraction of information from the context. For example, within our dataset, a declarative sentence like ``Only US citizens should be counted in the census'' was unequivocally identified by human annotators as reflecting an oppositional stance in the opinion task. Conversely, `System 2' tasks and stances often entail implied rather than overtly articulated stances. An illustrative case is a tweet that states, ``There are good people working at the Census Bureau with me, trying to count our citizens,'' which was classified by annotators as `Not Aware' of the citizenship question's inclusion, showing the subtlety with which stances can be embedded within the content.

\begin{figure}
    \centering
    \includegraphics[width=0.45\textwidth]{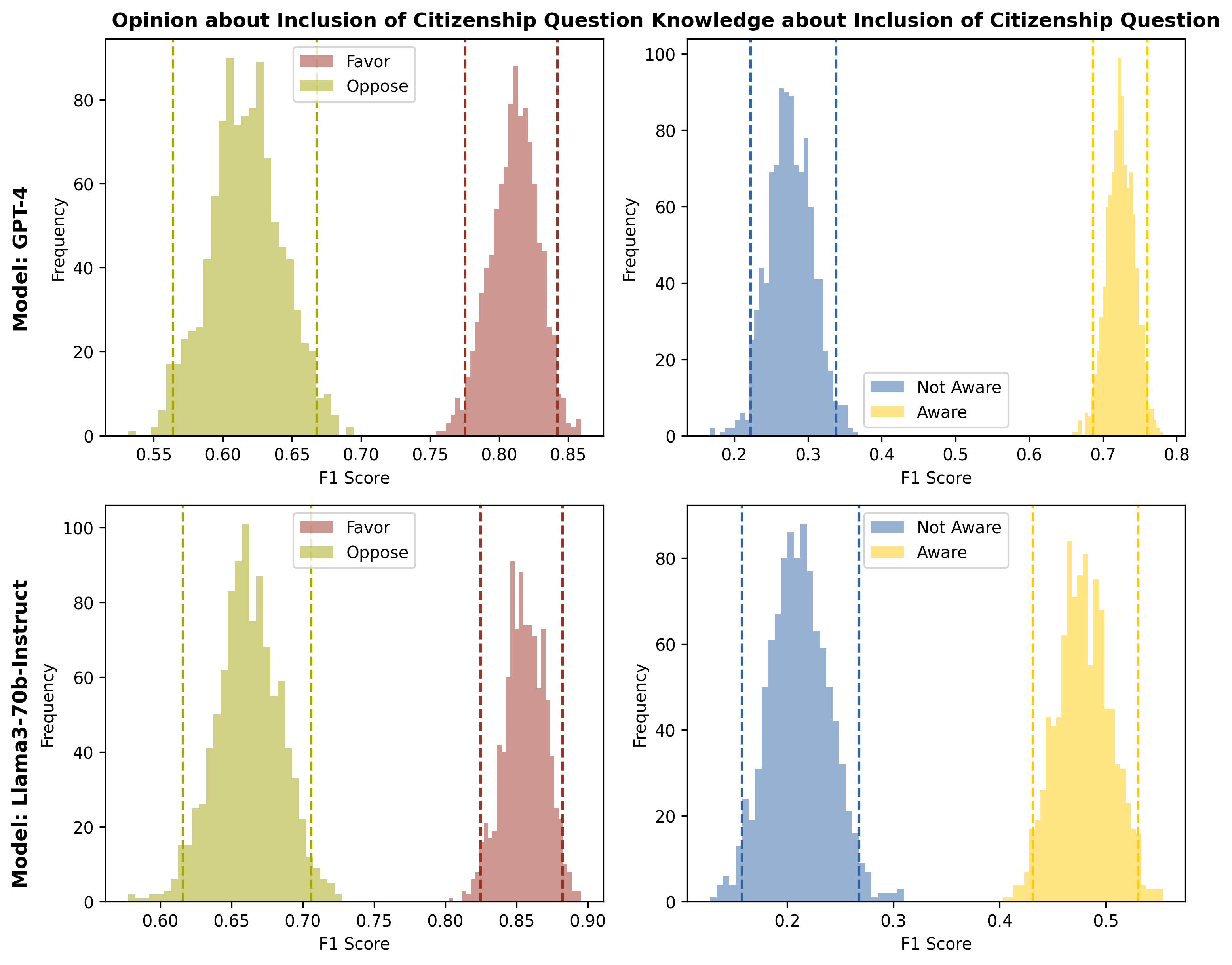}
    \caption{Bootstrapped F1 scores}
    \label{fig:fig1}
\end{figure}

As previously discussed, we adopted averaged human annotation as the benchmark and employed bootstrapping to estimate the confidence intervals of F1 scores across different questions and stances (scenarios), as can be seen in Table~\ref{tab:task_stance}. Figure \ref{fig:fig1} displays the performance of two LLMs, GPT-4 and Llama3-70b-Instruct, across four distinct scenarios, detailed in four separate panels. Each panel displays the F1 score distributions from bootstrapped samples for two categories of stances: ``Favor'' versus ``Oppose'' on one side, and ``Aware'' versus ``Not Aware'' on the other, with the top and bottom rows showing results from the two LLMs, respectively. The non-overlapping distributions for each category underscore a clear difference in model performance per task. These findings underscore that LLMs excel in opinion-based tasks but are less effective in  knowledge-based ones and encounter performance challenges detecting System 2 stances such as ``Oppose'' and ``Not Aware.'' This variation in performance highlights the impact of the required level of inferential reasoning on the congruence between LLM outputs and human judgments, particularly for `System 2' tasks. Moreover, our results affirm \textbf{H3}, illustrating that corpora typically comprise texts with varying degrees of explicitness, which, in turn, affects LLMs’ capability to accurately annotate or deduce human stances.

\section{Conclusion and Discussion}
Our investigation reveals that discrepancies between human annotators and LLMs predominantly arise when texts express stance less explicitly. That is to say, System 2 stance annotation, which requires more complex reasoning, poses more challenges for LLMs compared to the straightforward extraction of information in System 1 tasks. Another concern also raised that we don’t know much, but so far, it looks like LLMs struggle with cognitive tasks that are easy for people (\citealp{mitchell2023debate}; \citealp{dziri2024faith}). However, with the demonstration of how LLMs' inferential abilities have increased (e.g., from Llama2 to Llama3), we are optimistic that such discrepancies will narrow as LLMs continue to evolve.

We also demonstrated that while explicitness is traditionally viewed as an attribute inherent to individual texts, our analysis suggests that the characteristics of the corpus to which these texts belong might also influence whether stances are overtly expressed or subtly implied.

Considering these challenges, we propose specific strategies to optimize the use of LLMs in stance annotation tasks:
\begin{enumerate}
    \item Incorporate Prior Knowledge about the Corpus: Researchers should leverage their understanding of the corpus’s characteristics, informed by societal norms or topic sensitivity. Reading sample posts can confirm how explicit the texts are likely to be. This knowledge can predict whether stances will be explicitly stated or implied.

    \item Understand Human Inference in Annotation: annotators’ inferential processes when assigning stances are crucial to improving LLM performance of the analogous task. Identifying the specific skills human annotators apply by, for example, asking them to explain their thinking in a pop-up window, can help understand LLM decisions because, as we have shown, these often relate to human judgments.

    \item Conduct Pilot Annotations with LLMs and Craft Prompts Based on the outcomes: Running pilot annotations on a small sample with the chosen LLM can provide valuable insights. Requesting the rationale for each annotation decision can illuminate the LLM’s reasoning process. Insights from pilot annotations should guide the refinement of prompts to improve LLM accuracy. Understanding why an LLM might be confused by a text’s stance can inform adjustments to prompts or annotation guidelines. 
\end{enumerate}
By employing these and similar strategies, researchers may be able to better harness the potential of LLMs for stance annotation, accommodating the complexity and nuance of human language. By closely examining LLM performance and continuously refining annotation approaches, it is possible to align LLM decisions more closely with human judgment, thereby improving the reliability and validity of stance annotations.

In conclusion, LLMs perform well when human annotators do, and when LLMs fail, this often co-occurs with reduced human performance (agreement between annotators). When comparing LLMs with human annotators in stance detection tasks, it is essential to consider the nature of the text—whether stances are hidden (System 2) or easily extracted (System 1). LLMs are not off-the-shelf stance detection tools. To be used as effectively as possible for this purpose, researchers will craft prompts and models well suited to their specific corpora and help LLMs overcome some of the obstacles observed here.

\section{Limitations}

We acknowledge that our conclusions are based on a dataset comprising one knowledge question and one opinion question related to the 2020 US Census citizenship issue. Additionally, as we did not request Mturkers to provide reasons for their judgments, direct comparisons between human reasoning steps and those of LLMs are not feasible. When observing human judgments, researchers inferred the reasons behind human decisions during annotation, which are likely accurate but could be imprecise in some contexts.

Moreover, the eight LLMs used in this study do not represent the entire spectrum of currently available LLMs. However, it is noteworthy that we tested Llama2-70B-Chat on the same corpus and confirmed all three of our hypotheses. This bolsters our confidence that our findings will generalize to other LLMs, and we anticipate that LLMs will continue to improve with ongoing advancements.

\bibliography{custom}

\appendix
\section{Appendix}
\label{sec:appendix}

\subsection{Zero-Shot prompt template: SemEval and Opinion}

You have assumed the role of a human annotator. In this task, you will be presented with a tweet, delimited by triple backticks, concerning the \{topic\}. Please make the following assessment:

(1) Determine whether the tweet discusses the topic of the \{topic\}. If so, please indicate whether the Twitter user who posted the tweet favors, opposes, or has no opinion on the \{topic\}.

Your response should be only one option from those four: [favor, oppose, neutral, or irrelevant].

        Tweet: \{data\}

        Stance:

\subsection{Few-Shot prompt template: SemEval and Opinion}

You have assumed the role of a human annotator. In this task, you will be presented with a tweet, delimited by triple backticks, concerning the {topic}. Please make the following assessment:

(1) Determine whether the tweet discusses the topic of the \{topic\}. If so, please indicate whether the Twitter user who posted the tweet favors, opposes, or has no opinion on the \{topic\}.

Your response should be only one option from those four: [favor, oppose, neutral, or irrelevant].

For instance, if the tweet reads: \{example\}, your response should be: "\{example\_stance\}."

if the tweet reads: \{example2\}, your response should be: "\{example\_stance2\}."

        Tweet: \{data\}

\subsection{Zero-Shot prompt template: Knowledge}
You have assumed the role of a human annotator. In this task, you will be presented with a tweet, delimited by triple backticks, concerning the \{topic\}. Please make the following assessment:

(1) Determine whether the tweet discusses the topic of the \{topic\}. If so, please indicate whether the Twitter user who posted the tweet is aware of the fact that the Census Bureau will not include a question about citizenship?.

Your response should be formatted as follows: "Stance: [Yes, No, Can't tell, or Irrelevant]".

Thank you for your assistance with this task!
        Tweet: \{data\}
        
        Stance: 

\subsection{Few-Shot prompt template: Knowledge}

You have assumed the role of a human annotator. In this task, you will be presented with a tweet, delimited by triple backticks, concerning the {topic}. Please make the following assessment:

(1) Determine whether the tweet discusses the topic of the {topic}. If so, please indicate whether the Twitter user who posted the tweet is aware that the Census Bureau will not include a question about citizenship.

Your response should be formatted as follows: "Stance: [Yes, No, Can't tell, or Irrelevant]".

For instance, if the tweet reads: \{example\}, your response should be: "Stance: \{example\_stance\}."

if the tweet reads: \{example2\}, your response should be: "Stance: \{example\_stance2\}."

        Tweet: ```{data}```
        
        Stance: 
        
\section{Appendix}
\subsection{Output Parsing}
Our findings indicate that LLMs, depending on the model, varied in their ability to accurately interpret the same instruction. The expected response format was ``Stance: [Answer]''. However, many LLMs expanded their responses into comprehensive sentences, such as ``The tweet does not express a clear stance on the issue of the inclusion of a citizenship question...''. This inconsistency necessitated additional parsing of labels from the responses. We believe that prompt engineering could standardize these outputs, but to maintain consistency across all models, we refrained from customizing prompts for each LLM. Notably, ChatGPT models exhibited better alignment with the expected response format, compared with open-source LLMs, an aspect worth considering for text annotation tasks on a large scale, where reviewing every result individually is impractical. 

\end{document}